\documentclass[10pt,twocolumn,letterpaper]{article}

\usepackage{wacv}              

\usepackage{graphicx}
\usepackage{amsmath}
\usepackage{amssymb}
\usepackage{booktabs}
\usepackage[table]{xcolor}
\DeclareMathOperator*{\argmax}{argmax}
\usepackage{algorithm}
\usepackage{algpseudocode}

\usepackage{arydshln}
\usepackage[pagebackref,breaklinks,colorlinks]{hyperref}

\usepackage[capitalize]{cleveref}
\crefname{section}{Sec.}{Secs.}
\Crefname{section}{Section}{Sections}
\Crefname{table}{Table}{Tables}
\crefname{table}{Tab.}{Tabs.}

\def\wacvPaperID{2245} 
\def\confName{WACV}
\def\confYear{2025}

\begin{document}

\title{ACE: Action Concept Enhancement of Video-Language Models in Procedural Videos}

\author{Reza Ghoddoosian \hspace{1cm} Nakul Agarwal  \hspace{1cm} Isht Dwivedi \hspace{1cm} Behzad Darisuh  \\
Honda Research Institute, USA\\
{\tt\small \{reza\_ghoddoosian, nakul\_agarwal, idwivedi, bdariush\}@honda-ri.com}
}
\maketitle

\begin{abstract}

Vision-language models (VLMs) are capable of recognizing unseen actions. However, existing VLMs lack intrinsic understanding of procedural action concepts. Hence, they overfit to fixed labels and are not invariant to unseen action synonyms. To address this, we propose a simple fine-tuning technique, Action Concept Enhancement (ACE), to improve the robustness and concept understanding of VLMs in procedural action classification. ACE continually incorporates augmented  action synonyms and negatives in an auxiliary classification loss by stochastically replacing fixed labels during training. This creates new combinations of action labels over the course of fine-tuning and prevents overfitting to fixed action representations. We show the enhanced concept understanding of our VLM, by visualizing the alignment of encoded embeddings of unseen action synonyms in the embedding space. Our experiments on the ATA, IKEA and GTEA datasets demonstrate the efficacy of ACE in domains of cooking and assembly leading to significant improvements in zero-shot action classification while maintaining competitive performance on seen actions.
\end{abstract}


\section{Introduction}

Understanding human actions in procedural videos—such as cooking or assembly—has numerous applications, including training, human-robot interaction, and anomaly detection. Accurate understanding of anomalies and proficiency is critical enabling targeted interventions, as they can compromise safety, efficiency, and overall effectiveness. Anomalies can appear as missed steps, redundant actions, deviations from sequences, or departures from expert performance \cite{ata,lee2024error,prego}. Importantly, classification of previously unseen actions, as another form of anomaly, is essential for effective action recognition. For example, in a smart kitchen, it’s impractical or unsafe to gather data for scenarios like "cutting finger" or "spilling hot water," yet an intelligent assistant must identify and respond to such actions accurately and in real time.



\begin{figure}[t]
\centering
\includegraphics[width=0.48\textwidth,keepaspectratio]{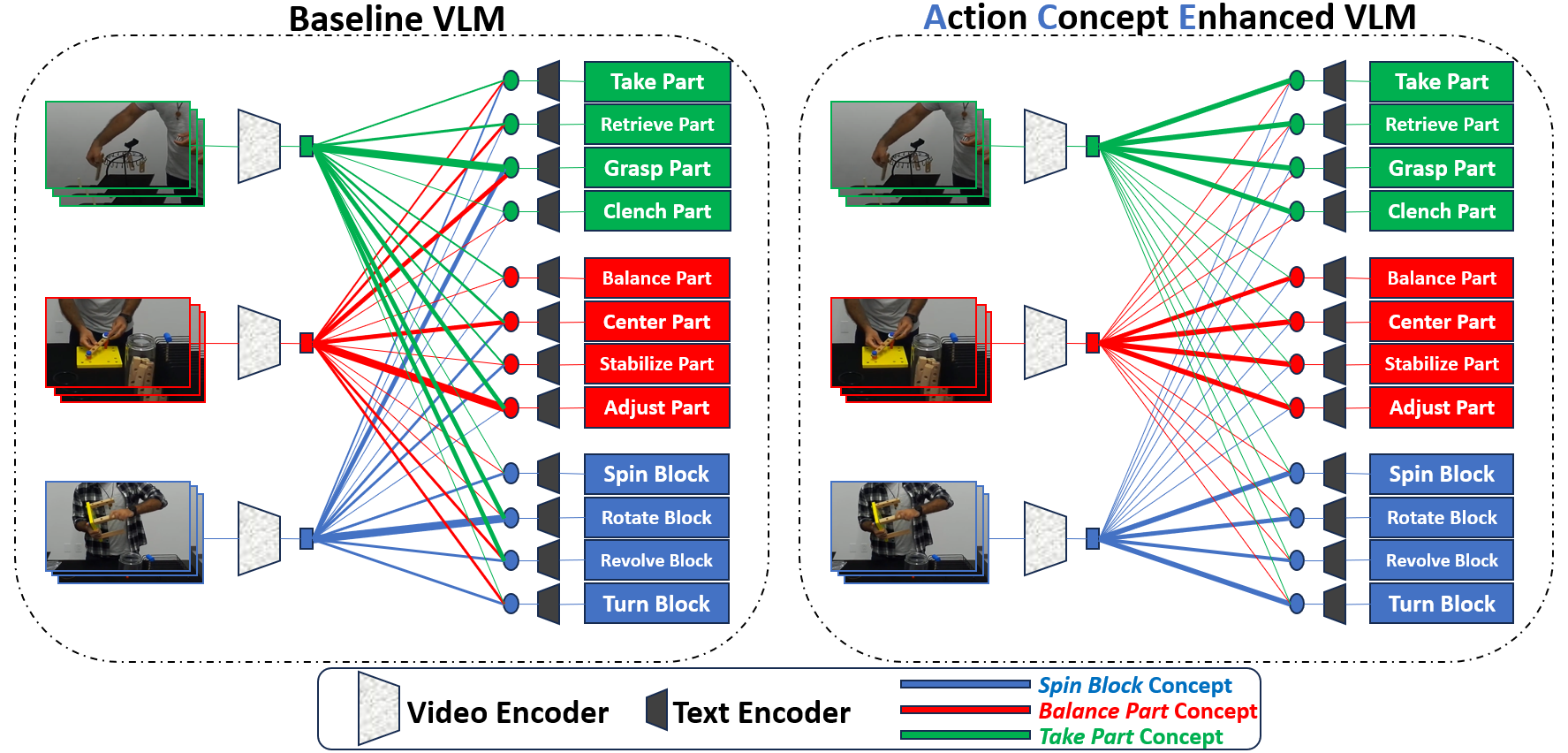}
\caption{Illustration of the similarity  between video and text representations for three action classes (concepts). Thicker lines indicate more similarity. Baseline VLMs (left) struggle with action synonym robustness. In contrast, ACE (right), improves accuracy in matching videos to action concepts, regardless of synonyms.}
\label{fig:concept}
\end{figure}

Vision-Language Models (VLMs) represent the state-of-the-art (SoTA) in zero-shot action recognition, where action categories are identified even if not explicitly seen during training. These models process videos and text through separate encoders, projecting them into a shared video-text embedding space. Here, a query video is matched to the closest text representation from unseen action labels manually curated by annotators. Since the actions and labels are unseen during training, VLMs must encode the broader concept of an action rather than the exact label. This enables the model to match a query video to its action class, regardless of the synonym used. Essentially, text representations describing the same action class should be projected close to each other in the embedding space. For example, as shown in Figure \ref{fig:concept}, a video of someone spinning a block should be associated with the relevant action class (denoted by the blue class), whether labeled "spin block," "rotate block," "revolve block," or "turn block."



Existing VLMs pretrained on large image-text datasets \cite{Text4Vis,BIKE,ViFiCLIP} often exhibit bias towards objects, failing to capture temporal action elements like verbs. Other VLMs \cite{proceVLR,lavilla}, pretrained on videos and internet transcripts, have text encoders that lack robustness, especially with fine-grained action synonyms in specialized and procedural domains. To address this, we propose a fine-tuning technique called Action Concept Enhancement (ACE) to improve VLM robustness and concept understanding. To our knowledge, we are the first to investigate action concept understanding in VLMs' and test their classification robustness against fine-grained and unseen action synonyms.


We leverage the knowledge of a Large Language Model (LLM) to construct a synonym tree, where each node is an action label and its descendants are synonyms. 
During training, we use a classification loss function where videos are classified into novel combinations of action synonyms and their negatives, randomly chosen from the tree. 
This method generates numerous action label combinations, ensuring the model encounters new or rare action sets each iteration, simulating classification into unseen categories.
The augmented synonyms introduce randomness and diversity, reducing overfitting to fixed verb representations, while negative labels help reduce bias toward objects. Our fine-tuning framework for VLMs integrates in-domain contextualization with the pretraining knowledge, enhancing recognition of unseen actions and understanding their concepts.


We evaluate our concept enhancement technique on the IKEA \cite{ikea}, ATA \cite{ata}, and GTEA \cite{gtea} datasets across cooking and assembly domains. Our method significantly outperforms the SoTA in recognizing unseen actions and understanding procedural concepts while performing competitively on seen actions.  The contributions of this paper are:

\begin{itemize} 
\setlength{\itemsep}{0pt}
\setlength{\parskip}{0pt}
\item To the best of our knowledge, we are the first to evaluate the action concept understanding of VLMs by testing their robustness to procedural and unseen action synonyms.
\item We introduce a fine-tuning mechanism that integrates in-domain knowledge into a pretrained model, enabling it to infer unseen procedural actions while maintaining performance on known actions. 
\item We use action synonyms stochastically during training to prevent VLMs from overfitting to fixed verbs and objects, leading to significant improvements in zero-shot action concept understanding. 
\item Our method, Action Concept Enhancement (ACE), is simple, generalizable, and easy to integrate into VLMs. We validate the classification efficacy of our method across different datasets and domains. 
\end{itemize}





\section{Related Work}
\label{sec:related}

\noindent \textbf{Zero-Shot Action Recognition (ZSAR).} ZSAR classifies videos into action categories not present in the training set. 
While transductive ZSAR uses test videos without labels during training, and generalized ZSAR handles both seen and unseen classes~\cite{gzsar,survey}, we adopt an inductive approach, evaluating unseen and seen classes separately without access to unseen videos or labels during training.
 Earlier approaches used word embeddings \cite{brattoli2020rethinking,piergiovanni2020learning} or manually annotated class attributes \cite{huynh2020fine} to represent action classes. Others decomposed text into fine-grained descriptions from the internet~\cite{qian2022rethinking,chen2021elaborative,wu2014zero} and encoded them with BERT~\cite{bert}. Methods like \cite{pu2022alignment,lin2022cross} generated unseen visual prototypes from linear combinations of seen ones, \cite{ReGen} used reinforcement learning for video captioning to classification, and \cite{obj1,obj2} utilized object priors for action inference. Recently, VLMs \cite{ViFiCLIP,ActionCLIP,VideoCLIP}, by aligning text and video embeddings  through contrastive learning, have outperformed earlier methods, showing strong generalization to unseen classes. Our work builds on VLMs for recognizing both seen and unseen procedural actions.

\noindent \textbf{Vision-Language Models for Action Recognition.} VLMs have been applied to action understanding tasks like action localization~\cite{detection1,stepformer,detection2,htstep}, alignment \cite{alignment}, and video retrieval \cite{retrieval,clover} in untrimmed videos. This paper focuses on step recognition in trimmed videos, where VLMs fall into two categories: image-based or video-based models. Image-based models, using a CLIP \cite{CLIP} encoder, leverage 400 million image-text pairs from the internet. These models adapt to video through prompt learning \cite{ju2022prompting,ActionCLIP}, adding temporal layers \cite{VerbsinAction,XCLIP}, or parameter-free fine-tuning on video datasets \cite{MAXI,ViFiCLIP,BIKE}. Despite large-scale pretraining, these methods struggle with knowledge retention and fail to capture temporal dynamics, focusing more on static objects than fine-grained action details like verbs. \cite{VerbsinAction} addresses this by generating hard negative captions from the SMiT dataset \cite{SMiT}, but improvements are mainly shown on Kinetics~\cite{Kinetics}, a dataset known for static bias.

Video-based models pretrain temporal encoders, like TimeSformer \cite{timesformer}, on large video datasets. Some models~\cite{hiervl,egovlp} pretrain on Ego4D~\cite{ego4d} for egocentric videos, while others \cite{VideoCLIP,distant,miech2020end,proceVLR,lavilla} use Howto100M \cite{howto100m} instructional videos with auto-transcribed narrations. These models capture temporal action dynamics, but their text encoders, like CLIP \cite{CLIP}, word2vec \cite{word2vec}, MPNet \cite{mpnet}, or BERT \cite{bert}, struggle with synonym variability for unseen actions. We propose a mechanism to improve robustness to label synonym variations. Additionally, while most VLMs focus on cross-dataset zero-shot inference, we evaluate zero-shot action recognition in a base-to-novel setting, fine-tuning the encoders on seen actions and testing on unseen ones in the same  dataset, as in \cite{ViFiCLIP}.

\noindent \textbf{Language Augmentation and  Concept Learning.} 
Augmenting language effectively distills more semantic knowledge into VLMs \cite{image_neg3,VerbsinAction,genre_classification,platypus}. In image understanding, methods like \cite{image_neg1,image_neg2,image_hier} add negative and hierarchical labels, while \cite{image_caption} uses stochastic captions to improve out-of-distribution image classification in CLIP models.
More relevant to our work are methods that augment labels for video action understanding. Recent efforts \cite{VerbsinAction,PAXION} focus on enhancing action knowledge by emphasizing verbs over objects. \cite{PAXION} uses tasks like video reversal and antonym detection to assess action knowledge, while \cite{VerbsinAction} generates hard negatives by replacing verbs in captions. Our approach reduces overfitting to both objects and verbs, improving concept understanding in VLMs, and is tested on robustness to unseen verb synonyms.


Generating labels from LLMs has improved skeleton-based action understanding \cite{skeleton-description} and self-supervised action recognition \cite{parellel_concept}. \cite{distant} uses WikiHow step descriptions to align text and video during pretraining. BIKE \cite{BIKE} recently employed a frozen CLIP encoder to extract relative attributes from a lexicon as a similarity measure for category labels. \cite{cap4video} and \cite{lavilla} use LLMs to generate auxiliary captions for retrieval and video representation learning in untrimmed videos. In contrast, our approach integrates stochastic synonym augmentation during fine-tuning for unseen action recognition in trimmed videos. Before VLMs, synonyms were used in ZSAR by \cite{alexiou2016exploring}, but their transductive method included test videos during training.

\section{Action Concept Enhancement (ACE)}
\label{sec:method}

\subsection{Problem Definition}
In training, our method processes a batch of size $B$ from trimmed procedural videos $\{I_n\}_{n=1}^B$ and their ground-truth action indices $\{y_n\}_{n=1}^B$. $y_n$ is the class index of the $n^{th}$ video, corresponding to one of the $C$ seen action categories $\boldsymbol{a}=\{a_i\}_{i=1}^C$. We define $\boldsymbol{a}$ as the \textit{default} or \textit{root} labels of seen action classes in the dataset. The goal is to fine-tune the pretrained vision encoder $\mathcal{E()}$ and text encoder $\mathcal{G()}$ so a trimmed test video is correctly classified into one of the action categories. This is achieved by aligning the query video embedding with the text embedding of its groundtruth action in the shared space.


We follow two separate classification scenarios at test time: first, classifying a test video into one of the seen classes $\boldsymbol{a}$; and second, classifying a test video into one of the previously unseen action labels $\acute{\boldsymbol{a}}=\{\acute{a}_i\}_{i=1}^{\acute{C}}$, regardless of the action synonyms used. This robustness is especially crucial for unseen actions, as the model has neither been optimized with nor expected any unseen action labels. Throughout this paper, bold notations distinguish sequences from single-element variables.

\subsection{Action Verb Synonym Trees}

\begin{figure}[t]
\centering
\includegraphics[width=0.48\textwidth,keepaspectratio]{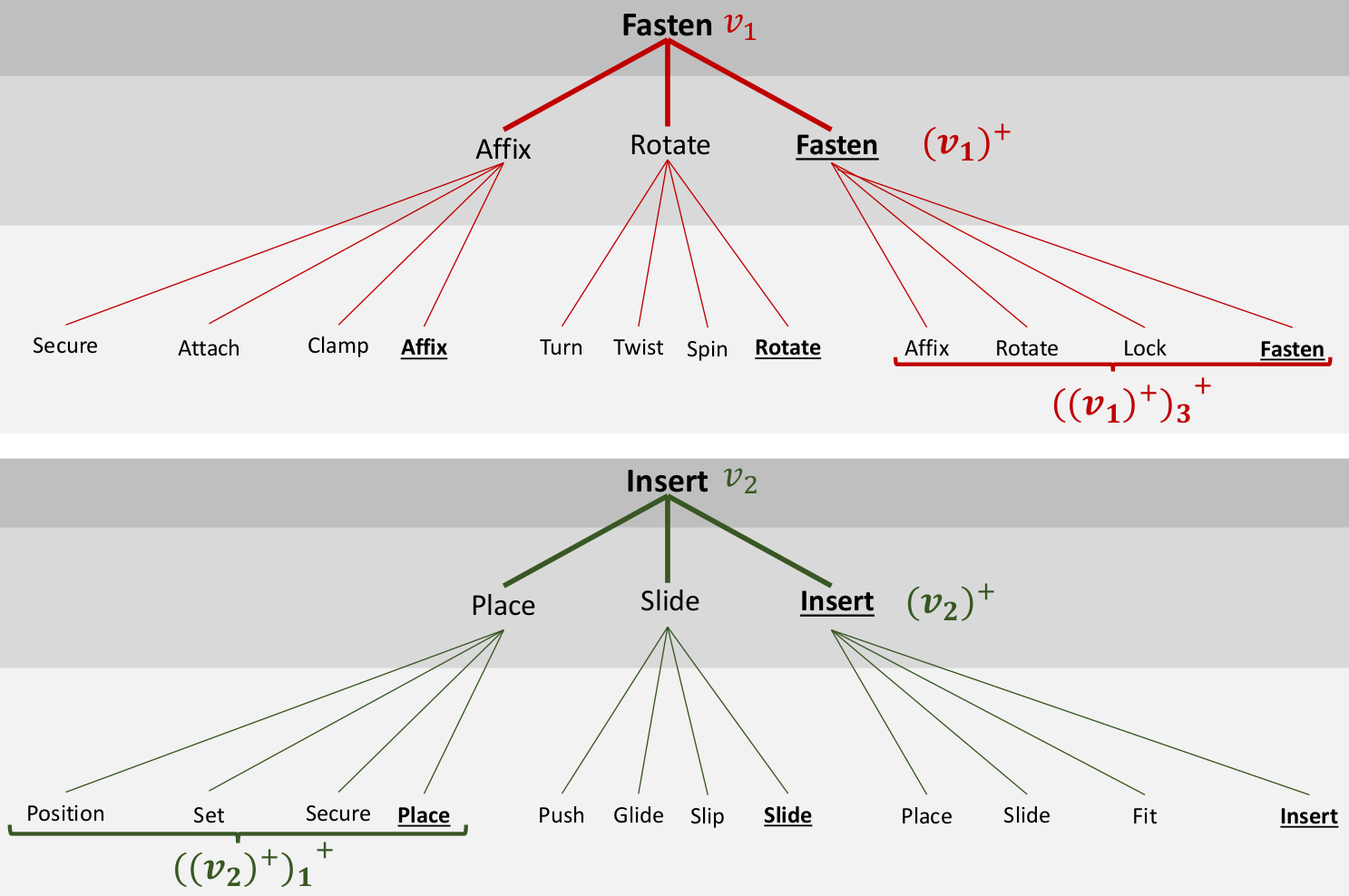}
\caption{Synonym trees for the action verbs 'fasten' and 'insert' and sample notations. Each tree represents an action concept, with replicated parent nodes highlighted in bold. Some second-order synonyms provide broader descriptions of the action.}
\label{fig:trees}
\end{figure}

We assume any procedural action $a$ can be decomposed into a verb $v$ and object\footnote{Without loss of generality, the object component can include multiple objects and prepositions.} $o$ pair, \ie, $a=v\oplus o$. 
We also define $\mathcal{V}(a)\rightarrow v$ and $\mathcal{O}(a)\rightarrow o$ as functions that map action $a$ to its corresponding verb and object components, respectively. Let $\boldsymbol{v}$ represent the set of $|\boldsymbol{v}|$ verb labels corresponding to root actions $\boldsymbol{a}$. As shown in Fig.\ref{fig:trees}, for each $v_i\in \boldsymbol{v}$, we establish a tree structure where $v_i$ is the root. In general, each parent node is a verb, and its $M$ children nodes are its synonyms, along with the parent verb itself. Each parent node is replicated as a child to ensure previous information is preserved at every semantic level. Concretely, children of node $v$ are denoted as $\boldsymbol{v^+}=\{{(v^+)}_i\}_{i=1}^M=Synonyms(v)\cup \{v\}$, where $\cup$ is the union operation, and synonyms are generated by an LLM. Although the number of children remains consistent at each level within a tree, it can vary across different levels.

In this paper, we build each tree up to second-order synonyms (\ie, synonyms of synonyms). However, in theory, these trees can extend to higher-order synonyms. Semantically, each tree corresponds to an action concept, and as trees deepen, action concepts overlap more and become less discriminative. This is because the connection between some higher-order synonyms and the root weakens, making action concepts coarser.


\subsection{Stochastic Action Concept Learning}
This section explains how the proposed synonym trees integrate into our learning framework. 

\noindent \textbf{Video-Label Alignment Loss.} In line with VLM training, the video encoder $\mathcal{E}()$ and text encoder $\mathcal{G}()$ map the input video $I_n$ and action labels $\boldsymbol{a}$ into a shared $D$-dimensional space. The cross-modal similarity $S(I_n,a_{y_n})$ between video $I_n$ and its groundtruth label $a_{y_n}\in\boldsymbol{a}$ is maximized, while the similarity of $I_n$ with other actions is minimized. The goal is to align related representations and separate unrelated ones. This alignment task is framed as a classification problem. For a batch of input data, the cross-entropy loss function $L_{fixed}$ maximizes $P(n,\boldsymbol{a})$, the probability of $I_n$ belonging to class $a_{y_n}$ given action labels $\boldsymbol{a}=\{a_i\}_{i=1}^C$.

\begin{small}
\begin{align}
P(n,\boldsymbol{a})={}&\frac{e^{S(I_n,a_{y_n})}}{\sum_{i=1}^{C}e^{S(I_n,a_{i})}} , \label{eq:P()}\\
L_{fixed}={}&-\frac{1}{B} \sum_{n=1}^{B}\log\big(P(n,\boldsymbol{a})\big) .\label{eq:loss1}
\end{align}
\end{small}
Here, the cross-modal similarity  $S(I,a)$ is defined as the average of cosine similarities between the video embedding and the text embeddings of $M$ children of action $a$:
\begin{small}
\begin{align}
S(I,a)=\frac{1}{\tau M} \sum_{i=1}^{M}<\mathcal{E}\big(I\big),\mathcal{G}\big((a^+)_i\big)> ,\label{eq:similarity}
\end{align}
\end{small}
where $\tau$ is the pre-defined temperature, $<\cdot,\cdot>$ indicates cosine similarity between two normalized embeddings, and $(a^+)_i=(\mathcal{V}(a)^+)_i \oplus \mathcal{O}(a)$. 
Computing similarity with an action via the average of its synonyms has three main advantages: first,  it brings related labels closer together through shared synonyms. Second, it helps to describe less familiar actions by their more recognizable synonyms.  Third, it simply adds more in-domain textual data for the model to learn from.

\noindent \textbf{Randomized Action Synonyms.} We further model our action concept enhancement as an auxiliary classification task where the pool of available action labels is randomly augmented from the set of known root actions $\boldsymbol{a}$.
Firstly, we define $\Tilde{x}$ as a sample randomly selected from the set $\boldsymbol{x}$. Accordingly, $\widetilde{\mathcal{V}(a_i)^+}$ refers to a verb randomly sampled from the synonyms of verb $\mathcal{V}(a_i)$  associated with action $a_i$.  Then, we leverage the verb synonym trees and Eq.\ref{eq:P()}, to extend $L_{fixed}$ by adding the auxiliary classification loss  $L_{rand}$ in Eq.\ref{eq:loss2}. Essentially, through $L_{rand}$, we categorize each video into one of the $C$ action classes labeled by a new set of randomized action synonyms $\boldsymbol{\widetilde{a^+}}$ at each training iteration. In detail, as specified below, $\boldsymbol{\widetilde{a^+}}$ is a random augmentation of seen action classes, where each action class is represented by its randomly chosen verb synonym:
\begin{small}
\begin{align}
\boldsymbol{\widetilde{a^+}}={}&\big\{\widetilde{\mathcal{V}(a_i)^+} \oplus \mathcal{O}(a_i)\big\}_{i=1}^C=\{(\widetilde{a_i^+})\}_{i=1}^C, \label{eq:action_positives} \\
L={}&-\frac{1}{B} \sum_{n=1}^{B}\log\big(P(n,\boldsymbol{a})\big)\underbrace{-\frac{1}{B} \sum_{n=1}^{B}\log\big(P(n,\boldsymbol{\widetilde{a^+}}}_{L_{rand}})\big).
\label{eq:loss2}
\end{align}
\end{small}

While for $L_{rand}$ a new set of randomized action synonyms $\boldsymbol{\widetilde{a^+}}$ is constructed per training iteration, $L_{fixed}$ uses the fixed root action labels throughout the entire training. Consequently, at every training iteration, each batch of videos is classified twice: once using the root labels and once using their randomized synonyms.

As root action labels are manually annotated in each dataset, they tend to be more precise descriptions of an action concept compared to AI-generated synonyms. Hence, the set of root labels in $L_{fixed}$ is fixed and serves as a reference point. This enables the video-language encoders learn the connection between root action labels and their synonyms within each action concept subspace. ``Concept subspace'' refers to the space covering the text representations of all synonyms associated with an action in the joint space.

Meanwhile, variable action labels in $L_{rand}$ prevent video-language encoders from overfitting to a single label, and instead learn different representations within a concept subspace. This enhances robustness to unseen action synonyms, and is beneficial in zero-shot recognition where actions and their labels are unknown. 
Our randomized augmentation technique can create up to $M^C$ different action label combinations which are rarely repeated during training. Effectively, this simulates test time classification, where videos are categorized into  unseen action labels.

Also, note that applying the similarity measure $S$ to first order synonyms in Eq.\ref{eq:loss2}, allows VLMs to learn action concepts based on second order synonyms of the tree.

\noindent \textbf{Shadow Negatives.}
We realized varying action synonyms through replacement of their verb components can bias the encoders to only  objects. In other words, encoders learn to align videos to their correct action labels by only focusing on the object component, which defeats the purpose of concept learning. In order to alleviate this limitation, we introduce \textit{shadow negative} as the $(C+1)^{th}$ category during classification. The shadow negative action shares the same object as the true action label; however, it pairs with a wrong verb. This approach compels the model to learn the verbs as well in order to accurately distinguish between the true label and its shadow negative.
Specifically, we utilize the verb synonym trees to define the pool of shadow negative verbs $\boldsymbol{\mathcal{V}(a_i)^{-}}$ associated with the root action $a_i\in\boldsymbol{a}$  as: 
\begin{small}
\begin{align}
\boldsymbol{\mathcal{V}(a_i)^{-}}=\bigcup_{j=1}^{C}\big(\boldsymbol{\mathcal{V}(a_j)^{+}}     \diagdown \boldsymbol{\mathcal{V}(a_i)^{+}}\big), \label{eq:verb_negatives}
\end{align}
\end{small}
where `$\diagdown$' refers to the set difference, \ie, children of $\mathcal{V}(a_j)$ that are not among the children of $\mathcal{V}(a_i)$.
At the beginning of each training iteration, for every class $i$, a shadow negative action $\widetilde{a_i^-}$  is constructed via random sampling from the pool of negative verbs $\boldsymbol{\mathcal{V}(a_i)^{-}}$ of that action: 
\begin{small}
\begin{align}
\widetilde{a_i^-}=\widetilde{{\mathcal{V}\small(a_i\small)}^-}\oplus\mathcal{O}(a_i).\label{eq:action_negatives}
\end{align}
\end{small}
Then, we update $P(n,\boldsymbol{a},\widetilde{a_{y_n}^-})$ as the probability of video $I_n$ belonging to class ${a_{y_n}}\in \boldsymbol{a}$ given the pool of positive action labels $\boldsymbol{a}$ and shadow negative $\widetilde{a_{y_n}^-}$. Adding the  shadow negative associated with the true action label of each video, extends the classification to $C+1$ classes:
\begin{small}
\begin{align}
P(n,\boldsymbol{a},\widetilde{a_{y_n}^-})=\frac{e^{S(I_n,a_{y_n})}}{\sum_{i=1}^{C}e^{S(I_n,a_{i})}+e^{S(I_n,\widetilde{a_{y_n}^-})}}.\label{eq:P2}
\end{align}
\end{small}
As a result, the final loss is also modified as follows:
\begin{small}
\begin{align}
L_f={}&-\frac{1}{B} \sum_{n=1}^{B}\Big(\underbrace{\log\big(P(n,\boldsymbol{a},\widetilde{a_{y_n}^-})\big)}_{L_{fixed}}+\underbrace{\log\big(P(n,\boldsymbol{\widetilde{a^+}},\widetilde{a_{y_n}^-})\big)}_{L_{rand}}\Big). \label{eq:loss_3}
\end{align}
\end{small}
\subsection{Training and Inference}
\noindent \textbf{Training:} In Alg. \ref{algorithm}, we summarize how to integrate ACE into the fine-tuning of VLMs for the task of video classification. We begin our algorithm by building the verb synonym trees $\big\{T\big(\mathcal{V}(a_i)\big)\big\}_{i=1}^C$. Next, at the beginning of each training iteration, as we process a batch, new randomized sets of action synonyms $\boldsymbol{\widetilde{a^+}}$ and shadow negatives $\boldsymbol{\widetilde{a^-}}$ are generated. These, along with root labels $\boldsymbol{a}$ and their respective children are encoded by the text encoder. Through Eq. \ref{eq:loss_3}, our algorithm then  engages each encoded video into two classification tasks involving $C+1$ categories. Consequently, this process encourages $\mathcal{E}$ and $\mathcal{G}$ encoders to explore action concepts  by stochastically aligning videos and synonyms within their corresponding concept subspace.

\begin{algorithm}
  \caption{Action Concept Enhancement (ACE)}
  \begin{algorithmic}[1]
    \Require{Input data $\mathcal{D}$ with videos and their groundtruth indices, the set of root action labels $\boldsymbol{a}$, and pretrained video and text encoders $\mathcal{E}$ and $\mathcal{G}$ respectively.}
    \Ensure{Fine-tuned video encoder $\mathcal{E}$ and text encoder $\mathcal{G}$ with enhanced  robustness to unseen action synonyms.}
    \State $\big\{T\big(\mathcal{V}(a_i)\big)\big\}_{i=1}^C\leftarrow\text{LLM}(\boldsymbol{a})$ \Comment{Verb synonym tree $T\big(\mathcal{V}(a_i)\big)$ rooted in action verb $\mathcal{V}(a_i)$ for $a_i\in \boldsymbol{a}$.}
    
    \For{every epoch }:
       \For{ batch $\{I_n,y_n\}_{n=1}^B\gets\mathcal{D}$}:
        \State $\boldsymbol{\widetilde{a^+}}=\big\{ \widetilde{a_i^+}\gets T\big(\mathcal{V}(a_i)\big)\big\}_{i=1}^C$ \Comment{Eq.\ref{eq:action_positives}}
        \State  $\boldsymbol{\widetilde{a^-}}=\Big\{ \widetilde{a_i^-}\gets\big\{ T\big(\mathcal{V}(a_j)\big)\big\}_{j=1}^C\Big\}_{i=1}^C$ \Comment{Eq.\ref{eq:action_negatives}}
         \State $L_f\gets$ Use Eq. \ref{eq:loss_3} to calculate  $L_{fixed}$ and $L_{rand}$
         \State Backprop and optimize $\mathcal{E}$ and $\mathcal{G}$ 
       \EndFor
    \EndFor \\
\Return Action Concept Enhanced (ACEd) $\mathcal{E}$ and $\mathcal{G}$
  \end{algorithmic}\label{algorithm}
\end{algorithm}

\noindent \textbf{Inference:} During inference, we classify query video $I_n$ into the action class that has the highest similarity $S$ with the query video, \ie, $\argmax_{a\in \mathbb{A}} S(I_n,a)$. Following \cite{ViFiCLIP}, inference is done in two separate modes of \textit{base} and \textit{novel}, where $\mathbb{A}$ is the set of known classes $\boldsymbol{a}$ in the base mode and the set of unseen classes $\acute{\boldsymbol{a}}$ in the novel mode.  In addition, in both base and novel modes, synonym trees are constructed, so $\mathbb{A}$ can be represented by the root action labels or the synonyms of the root labels. 
Note, we do not use any shadow negatives during inference. 

\section{Experiments}
\label{sec:experiments}

\subsection{Experimental Setup}

\begin{table*}[t]
\footnotesize\setlength{\tabcolsep}{2.0 pt}
\centering
\caption{Procedural action classification on 2 exocentric datasets. HM is the Harmonic Mean of seen and unseen results \cite{ViFiCLIP}. SRT measures the robustness of VLMs to unseen action synonyms via mean$\pm$std. Results follow the `\{\textit{acc}\}/\{\textit{F1}\}' format with 2nd bests underlined.}
\begin{tabular}{c|c|c c c|c|c c c|c}
\hline
\multicolumn{2}{c|}{}& 
\multicolumn{4}{|c|}{ATA Dataset \textit{[10 base and 5 novel classes]}} & \multicolumn{4}{|c}{IKEA Dataset \textit{[21 base and 10 novel classes]}} \\ 
\cline{3-10}
 \multicolumn{2}{c|}{} & \multicolumn{3}{|c|}{Default Labels} & SRT & \multicolumn{3}{|c|}{Default Labels} & SRT \\ 
\hline
 Method &Pretraining Videos& Seen & Unseen & HM & Unseen & Seen & Unseen &HM & Unseen \\ 
\hline
Random & - & \cellcolor{gray!20}- & 20.0/18.5 &\cellcolor{gray!20}- & 20.0$\pm$0/18.5$\pm$0 & \cellcolor{gray!20}- & 10.0/7.1 & \cellcolor{gray!20}-  & 10.0$\pm$0/7.1$\pm$0 \\
ViFi (ft)\cite{ViFiCLIP} & Just Images & \cellcolor{gray!20}\underline{93.1}/\underline{94.2} & 33.5/25.4 &\cellcolor{gray!20}49.3/40.0 & \underline{44.8}$\pm$5.4/\underline{33.2}$\pm$3.7 & \cellcolor{gray!20}78.7/61.0 & \underline{44.3}/29.6 & \cellcolor{gray!20}\underline{56.7}/39.9 & \underline{34.2}$\pm$7.5/27.6$\pm$4.6  \\
ViFi (pr)\cite{ViFiCLIP} & ASM101\cite{ASM101} &\cellcolor{gray!20}88.5/91.2 & \underline{41.1}/29.0 & \cellcolor{gray!20}\underline{56.1}/44.0 & 33.4$\pm$7.4/27.2$\pm$3.5 & \cellcolor{gray!20}73.8/56.3 & 36.1/20.5 & \cellcolor{gray!20}48.5/30.1 & 32.5$\pm$3.8/23.1$\pm$3.4  \\
BIKE\cite{BIKE} & K400\cite{Kinetics} & \cellcolor{gray!20}93.0/92.2 & 29.8/18.8 &\cellcolor{gray!20}45.1/31.2 & 25.6$\pm$11.7/17.7$\pm$6.9& \cellcolor{gray!20}77.2/63.7& 36.5/31.3 & \cellcolor{gray!20}52.2/43.3 & 27.3$\pm$10.9/30.1$\pm$7.3 \\
Text4Vis\cite{Text4Vis}& K400\cite{Kinetics} & \cellcolor{gray!20}\textbf{93.5/95.0} & 38.7/30.4 &\cellcolor{gray!20}54.7/46.1& 42.7$\pm$8.6/26.0$\pm$6.7 & \cellcolor{gray!20}\textbf{87.6/71.6} & 39.3/\underline{39.4} & \cellcolor{gray!20}54.3/\underline{50.8} & 22.8$\pm$10.7/\underline{33.1}$\pm$5.8  \\
ProcVLR\cite{proceVLR} & Howto100M\cite{howto100m} & \cellcolor{gray!20}89.8/90.1 & 37.1/\underline{31.3} &\cellcolor{gray!20}52.5/\underline{46.4}& 43.8$\pm$10.6/33.0$\pm$6.9 & \cellcolor{gray!20}82.3/62.5 & 37.0/30.3 & \cellcolor{gray!20}51.0/40.8 & 32.1$\pm$13.6/27.1$\pm$5.5 \\
\hline
\textbf{Ours} & Howto100M\cite{howto100m} & \cellcolor{gray!20}90.0/91.7 & \textbf{60.8/47.3} &\cellcolor{gray!20}\textbf{72.6/62.4}& \textbf{59.4$\pm$3.2/43.3$\pm$4.2} & \cellcolor{gray!20}\underline{82.8}/\underline{63.9}& \textbf{54.5/45.5} & \cellcolor{gray!20}\textbf{65.7/53.2} & \textbf{45.9$\pm$6.1/41.1$\pm$4.2}  \\ \hline
\end{tabular}\label{table:comparison}
\end{table*}

\noindent\textbf{Datasets.}
We assess the efficacy of ACE in two procedural domains, cooking and assembly, using the following three \textit{real-time} datasets: 1) \textit{ATA} \cite{ata} is a toy assembly dataset with 12k trimmed videos and 15 action classes, recorded by 4 exocentric cameras. ATA's training and test splits consists of 27 and 4 participants, respectively.  2) \textit{IKEA}\cite{ikea} features table and drawer assemblies from 3 exocentric viewpoints, with 16k trimmed videos and 31 action classes. It’s divided into 5 splits based on the environment, and results are averaged across these splits unless noted otherwise.
3) \textit{GTEA}\cite{gtea} is an egocentric dataset where 4 subjects prepare 7 different dishes. It comprises  525 videos and 10 action verb classes. Following \cite{mstcn}, we perform classification on the verb classes in order to challenge zero-shot classification methods that are biased to objects and ignore verbs. We use 4 fold cross-validation with a subject left out for testing each time.

\noindent\textbf{Evaluation Protocol.}
In each dataset, we select one-third of  action classes with the least frequent verb  as the set of unseen \textit{(novel)} actions for zero-shot classification. The remaining classes are used as the seen \textit{(base)} actions. This strategy increases the chance that novel and base verb sets are also mutually exclusive. The base actions in the training split are used for fine-tuning, while the base and novel actions in the test set are used to evaluate seen and unseen action recognition, respectively. 

\noindent\textbf{Evaluation Metrics.}
We report classification results using Top 1 Accuracy (\textit{acc}) and \textit{F1} score. While \textit{acc} is averaged over all videos and is the most commonly used metric\cite{ViFiCLIP}, it is  skewed towards classes with more samples. In contrast, \textit{F1} is computed as the average F1 score of all classes in the test split and weighs classes equally.
Additionally, in order to test concept understanding and robustness to unseen action synonyms, 
we show the mean and standard deviation (std) over 10 different test runs. In each run,  we repeat the classification of test videos, using a new combination of randomly-selected unseen action synonyms. Hence, we refer to these experiments as the Synonym Robustness Test (SRT). To ensure fairness, SRT for all methods is based on the same sets of action synonyms. We provide our generated sets of unseen action synonyms in the supp. material.

\noindent\textbf{Implementation Detail.}
We use a 12-layer TimeSformer \cite{timesformer} video encoder pretrained on Howto100M via ProcVLR \cite{proceVLR} and the original 12-layer CLIP text encoder (ViT-B/16). GPT-4 \cite{gpt4} generates synonyms for our method.  The number of first and second order children in synonym trees are 2,9 and 11 for the IKEA, GTEA and ATA datasets, respectively. Furthermore, batch size is set to 16, temperature $\tau$ is adjusted to 0.02, and SGD optimizes the model for up to 15 epochs. Refer to the supp. material for more details. We intend to release our code and parameters publicly.

\subsection{Comparison with Baselines}

\noindent\textbf{Baselines.} 1) \textit{Random} guess accuracy and F1 score, which is calculated considering the label distribution and equal guessing probability for each unseen action. This provides context to zero-shot prediction of other methods. 2) \textit{ViFi (ft)} \cite{ViFiCLIP} fine-tunes non-frozen CLIP image and text encoders using averaged video frame encodings. 3) \textit{ViFi (pr)}  is a variation of \cite{ViFiCLIP}, where image and text encoders are pretrained on procedural videos of the ASM101 dataset \cite{Kinetics} and are kept frozen afterwards. Instead, learnable prompting layers are integrated in both encoders and trained during fine-tuning. 4) \textit{BIKE} \cite{BIKE} represents methods that utilize text augmentation. It generates 200 extra language attributes by GPT-4 for training and inference. BIKE pretrains CLIP vision and text encoders separately on Kinetics-400 (K400) \cite{Kinetics}. 5) \textit{Text4Vis} \cite{Text4Vis} also uses CLIP encoders and pretrains on the K400 dataset while keeping the text encoder fixed as a classifier. 6) \textit{ProcVLR} \cite{proceVLR}, similar to us, pretrains on Howto100M instructional videos and fine-tunes TimeSformer to encode videos. However, their CLIP text encoder is kept frozen.

 We compare with these SoTA VLMs based on the availability of their codes and pretrained model checkpoints. Results are reported after running authors' source code on our procedural video datasets.


\begin{table}[t]
\begin{center}
\footnotesize\setlength{\tabcolsep}{2.5 pt}
\caption {Action classification of the egocentric GTEA videos.\label{table:comparison_gtea}} 
\begin{tabular}{c|c c c|c}
\hline
\multicolumn{1}{c|}{}& 
\multicolumn{4}{|c}{GTEA Dataset \textit{[6 base and 4 novel classes]}}   \\ 
\cline{2-5}
 \multicolumn{1}{c|}{} & \multicolumn{3}{|c|}{Default Labels} & SRT \\ 
\hline
 Method & Seen & Unseen & HM & Unseen \\ 
\hline
Random & \cellcolor{gray!20}- & 25/22.8 &\cellcolor{gray!20}- & 25.0$\pm$0/22.8$\pm$0 \\
ViFi (ft)\cite{ViFiCLIP} & \cellcolor{gray!20}75.9/72.8 & 27.5/21.1 & \cellcolor{gray!20}40.4/32.7 &33.1$\pm$15.5/20.8$\pm$14.0 \\
ViFi (pr)\cite{ViFiCLIP} & \cellcolor{gray!20}58.4/50.2 & 25.6/13.1 & \cellcolor{gray!20}35.6/20.8 &24.4$\pm$20.7/15.0$\pm$14.9  \\
BIKE\cite{BIKE} & \cellcolor{gray!20}63.8/62.5& 50.0/33.9& \cellcolor{gray!20} 56.1/ 43.9 &39.9$\pm$30.6/29.7$\pm$12.7 \\
Text4Vis\cite{Text4Vis}& \cellcolor{gray!20}\underline{82.2}/\underline{81.3} & \underline{63.8}/\textbf{47.4} & \cellcolor{gray!20}\underline{71.8}/\textbf{59.8} &40.0$\pm$27.4/\underline{30.3}$\pm$22.3  \\
ProcVLR\cite{proceVLR} & \cellcolor{gray!20}68.0/64.2 & 50.3/36.1 & \cellcolor{gray!20}57.8/46.2 &\underline{45.0}$\pm$16.9/28.8$\pm$15.5\\
\hline
\textbf{Ours}  & \cellcolor{gray!20}\textbf{85.1/84.4} & \textbf{67.2}/\underline{41.0} &\cellcolor{gray!20}\textbf{75.1}/\underline{55.2}& \textbf{45.0$\pm$16.8/32.4$\pm$14.1}\\ \hline
\end{tabular}
\end{center}
\end{table}

\noindent\textbf{Procedural Action Classification.}
Table \ref{table:comparison} and \ref {table:comparison_gtea} compare our classification results with  existing VLMs on exo and ego datasets, respectively. When tested against varying synonym labels, through SRT, our method shows the most robust performance. It achieves the highest mean $\textit{acc}$ and $\textit{F1}$ while maintaining low std in all datasets. GTEA test splits include fewer videos, which explains the higher overall std for all methods. While image-based models like ViFi also show low std for varying verbs, this is largely due to overfitting to objects, resulting in significantly lower mean values.

For the sake of completeness, we also compare results using default action labels. Our method significantly beats the SoTA  in zero-shot classification for 5 out of 6 metrics across three datasets while remaining competitive on seen classes.  
Although Text4Vis\cite{Text4Vis} classifies seen actions more accurately in 2 of the 3 datasets, it doesn't generalize well to unseen actions, leading to lower overall performance (HM) compared to us. \cite{Text4Vis}'s success with base classes is mainly due to fine-tuning all encoder layers, whereas we only fine-tune the last three. However, as shown in Fig.\ref{fig:finetuning}, fine-tuning more layers can further improve our seen action classification too.
In general, the HM score reflects the trade-off between base and novel classes, and ACE improves the HM $\textit{acc}$ of the second best baselines by up to 16\%, 11\% and 4\% for ATA, IKEA  and GTEA datasets, respectively.
Importantly, ProcVLR \cite{proceVLR} is our one-to-one competitor as we share the same backbone and pretraining dataset. ACE enhances the action concept understanding of \cite{proceVLR} consistently on all datasets for both base and novel actions.

\subsection{Ablation Study and Fine-Tuning Analysis}
We ablate our zero-shot Action Concept Enhancement based on the statistics over the 10 experiments of the Synonym Robustness Test. Results shown as `\{\textit{acc}\}/\{\textit{F1}\}'.
\begin{table}[t]
\begin{center}
\footnotesize\setlength{\tabcolsep}{3.0 pt}
\caption {Impact of different modules of ACE on unseen actions.} .\label{table:system_contribution}
\begin{tabular}{l|c c |c c}
\hline
 \multicolumn{1}{c|}{} & \multicolumn{2}{|c|}{ATA Dataset} &\multicolumn{2}{|c}{IKEA Dataset} \\ 
\hline
 \multicolumn{1}{c|}{Setting} & \textit{acc} & \textit{F1} & \textit{acc}& \textit{F1} \\ 
\hline
w/o leaf augmentation & 45.2$\pm$11.6& 33.1$\pm$8.1 & 34.0$\pm$9.9&33.1$\pm$6.4 \\
w/o shadow negatives & 56.4$\pm$2.9 & 36.8$\pm$3.0 & 35.4$\pm$9.3&36.3$\pm$4.0  \\
w/o $L_{rand}$ & 41.1$\pm$7.6 & 29.5$\pm$5.5 & 44.4$\pm$10.6 &36.2$\pm$4.2 \\
w/o $L_{fixed}$& 57.7$\pm$3.5 & 42.5$\pm$4.5 & 40.0$\pm$7.5 &36.3$\pm$5.0  \\
\hline
 \textbf{ACE}  & \textbf{59.4$\pm$3.2} & \textbf{43.3$\pm$4.2}&\textbf{45.9$\pm$6.1}& \textbf{41.1$\pm$4.3}\\ \hline
\end{tabular}
\end{center}
\end{table}

\noindent\textbf{Ablation of ACE Components.}
Table \ref{table:system_contribution} highlights how the zero-shot performance of ACE drops whenever each component of the algorithm is removed.
We use the term ``leaf augmentation'' to describe representation of an action by the average of its synonym in the similarity measure of Eq.\ref{eq:similarity}. Leaf augmentation has the most significant impact on robustness, as it greatly reduces standard deviation.
Also, Excluding shadow negatives leads to overfitting to objects, which reduces result variability but at the cost of lower overall average performance. This drop is less pronounced in the ATA dataset where unseen classes are easier to identify through objects alone. In contrast,  the IKEA dataset involves multiple unseen actions that share the same objects, making action verb comprehension more crucial.
Addition of $L_{rand}$ also shows that the performance gain is not only due to extra text data in leaf augmentation, and the stochastic selection of synonyms plays a key role to prevent overfitting to fixed action verbs.

\noindent\textbf{Sensitivity to the Quality of Action Synonyms.}
In order to evaluate how important the quality of generated synonyms are, an expert human annotator rebuilds the synonym trees by manually modifying the GPT-generated ones. The resulting manual synonyms fit the context of cooking better and there is no shared first order synonyms across different concepts. Manual synonym trees are built for both base and novel root actions in training and inference, respectively. We chose the GTEA dataset for this experiment (Table \ref{table:synonym_quality}) as its scale makes the manual annotation plausible. As expected, better quality of manually-annotated synonyms improves the robustness of the model with the
best result achieved when the model is trained and tested on manually-annotated synonym trees. Note, the mismatch between the manually-annotated  synonyms during training and the GPT-generated synonyms during inference can negatively affect the model. Despite this, ACE still outperforms SoTA in zero-shot classification with GPT-generated synonyms. During SRT, the same randomization seeds are used for both trained models.

\begin{table}[t]
\begin{center}
\footnotesize\setlength{\tabcolsep}{3.0 pt}
\caption {Comparison of ACE when action synonyms are generated by GPT-4 vs. a human annotator on the GTEA dataset.} .\label{table:synonym_quality}
\begin{tabular}{l|c|c}
\hline
 \multicolumn{1}{c|}{\textit{M=9}} & \multicolumn{1}{|c|}{GPT Test Tree} &\multicolumn{1}{|c}{Manual Test Tree} \\ 
\hline 
GPT Training Tree & 45.0$\pm$16.8/32.4$\pm$14.1& 56.3$\pm$13.9/44.3$\pm$17.7\\
\hline
Manual Training Tree & 41.1$\pm$16.6/28.2$\pm$13.4& \textbf{63.6$\pm$16.4/45.4$\pm$15.3}
\end{tabular}
\end{center}
\end{table}

\begin{figure}[t]
\centering
\includegraphics[width=0.48\textwidth,keepaspectratio]{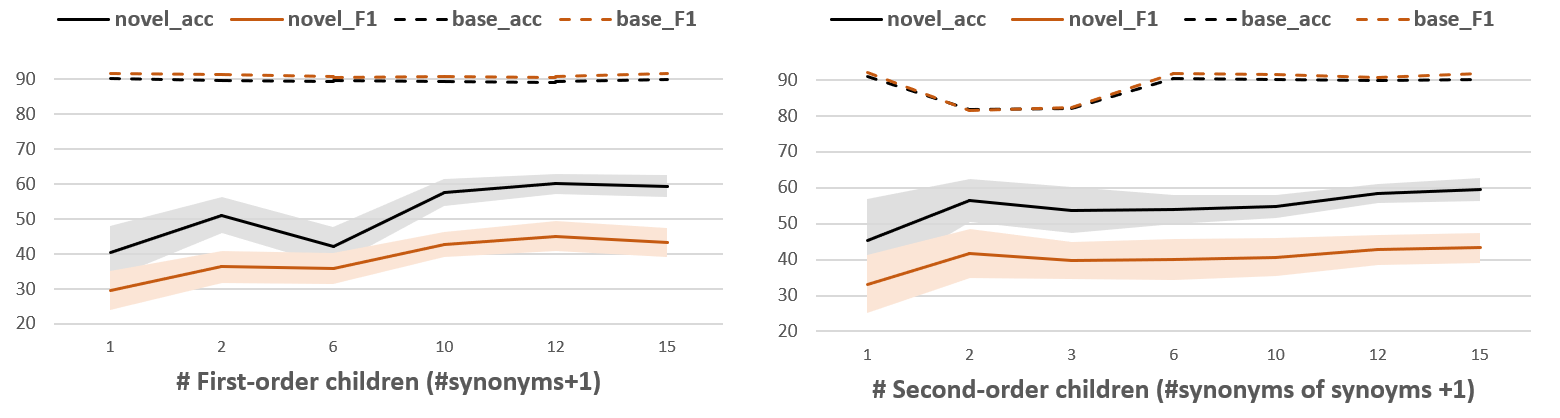}
\caption{Impact of the quantity of augmented synonyms on mean and std (shaded area) for novel and base actions of the ATA dataset.}
\label{fig:syns_sensitivuty}
\end{figure}

\noindent\textbf{Sensitivity to the Number of Action Synonyms.}
Fig. \ref{fig:syns_sensitivuty} illustrates ACE's sensitivity to the number of first and second order children in the synonym trees. 
Monotonically non-increasing results on the ATA dataset show that more synonyms does not necessarily lead to better results. This is because the quality of synonyms is also a deciding factor (Table. \ref{table:synonym_quality}), especially  when the number of synonyms is small. However, evidently, using any number of added first or second order synonyms improves the VLM robustness compared to when no synonym is augmented. This impact is less significant on seen actions. Eventually, results tend to converge as the number of synonyms increases.

\begin{figure}[t]
\centering
\includegraphics[width=0.48\textwidth,keepaspectratio]{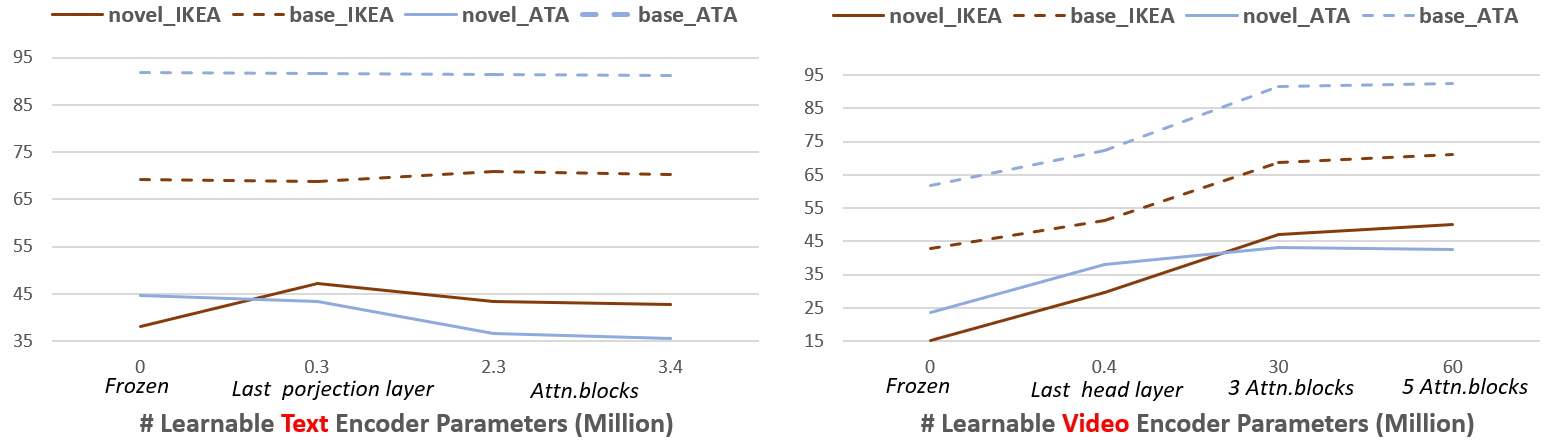}
\caption{Impact of fine-tuning various layers of video-text encoders on the mean F1 score. Results on ATA and split 1 of IKEA.}
\label{fig:finetuning}
\end{figure}

\noindent\textbf{How Many Encoder Layers to Fine-Tune?}
We study the extent to which each encoder should be fine-tuned in order to adjust the pretraining knowledge to new action concepts without losing prior information. Specifically, F1 scores in Fig.\ref{fig:finetuning}-left show that deep finetuning of text encoder has minimal effect on the base actions, but degrades the zero-shot performance as the text encoder starts forgetting some of its pre-acquired knowledge. On the other hand, the video encoder (TimeSformer) can benefit from deeper fine-tuning (Fig.\ref{fig:finetuning}-right). Not only this boosts results on seen actions, but also sometimes leads to better zero-shot performance as observed for the IKEA dataset. Nevertheless, fine-tuning more layers is computationally expensive and zero-shot performance gain is not always guaranteed on less diverse datasets such as ATA. Hence, we found the best trade-off to be training only the last three attention blocks of TimeSformer and the final projection layer of the text encoder.

\noindent\textbf{Pretrained vs. In-Domain Knowledge.}
Table \ref{table:finetuneVSpretraining} shows that ACE effectively adjusts the prior knowledge of the pretrained TimeSformer by incorporating new in-domain information. Otherwise, applying ACE to a randomly-initialized TimeSfomer is not sufficient to learn action embeddings on our datasets. 
Moreover, although the pretrained TimeSformer excels in zero-shot performance on YouTube-based datasets like COIN \cite{COIN,proceVLR}, it performs poorly on our real-time and unedited benchmarks without further fine-tuning. This is due to a domain shift, as TimeSformer was trained on YouTube procedural videos.
\begin{table}[t]
\begin{center}
\footnotesize\setlength{\tabcolsep}{3.0 pt}
\caption {Impact of pretraining the TimeSformer on Howto100M before ACE fine-tuning. Random initialization used otherwise.} .\label{table:finetuneVSpretraining}
\begin{tabular}{c c|c c|c c}
\hline
\multicolumn{2}{c|}{Setting} & \multicolumn{2}{|c|}{ATA}& \multicolumn{2}{|c}{IKEA-split1} \\
\hline
\multicolumn{1}{c}{Pretraining} & \multicolumn{1}{c|}{ACE} &\multicolumn{1}{|c|}{Seen}  & \multicolumn{1}{|c|}{Unseen}  & \multicolumn{1}{|c|}{Seen}  & \multicolumn{1}{|c}{Unseen} \\ 
\hline 
\checkmark & $\times$ & - & 18.8/10.8 &-&6.0/5.0\\
$\times$ & \checkmark & 24.6/9.7 & 20.5/13.2 &21.9/1.7&34.0/8.0	\\
\checkmark & \checkmark & 81.8/81.5 & 56.4/41.6 &88.0/71.1&58.6/50.2	\\
\end{tabular}
\end{center}
\end{table}

\begin{figure*}[t]
\centering
\includegraphics[width=0.95\textwidth,keepaspectratio]{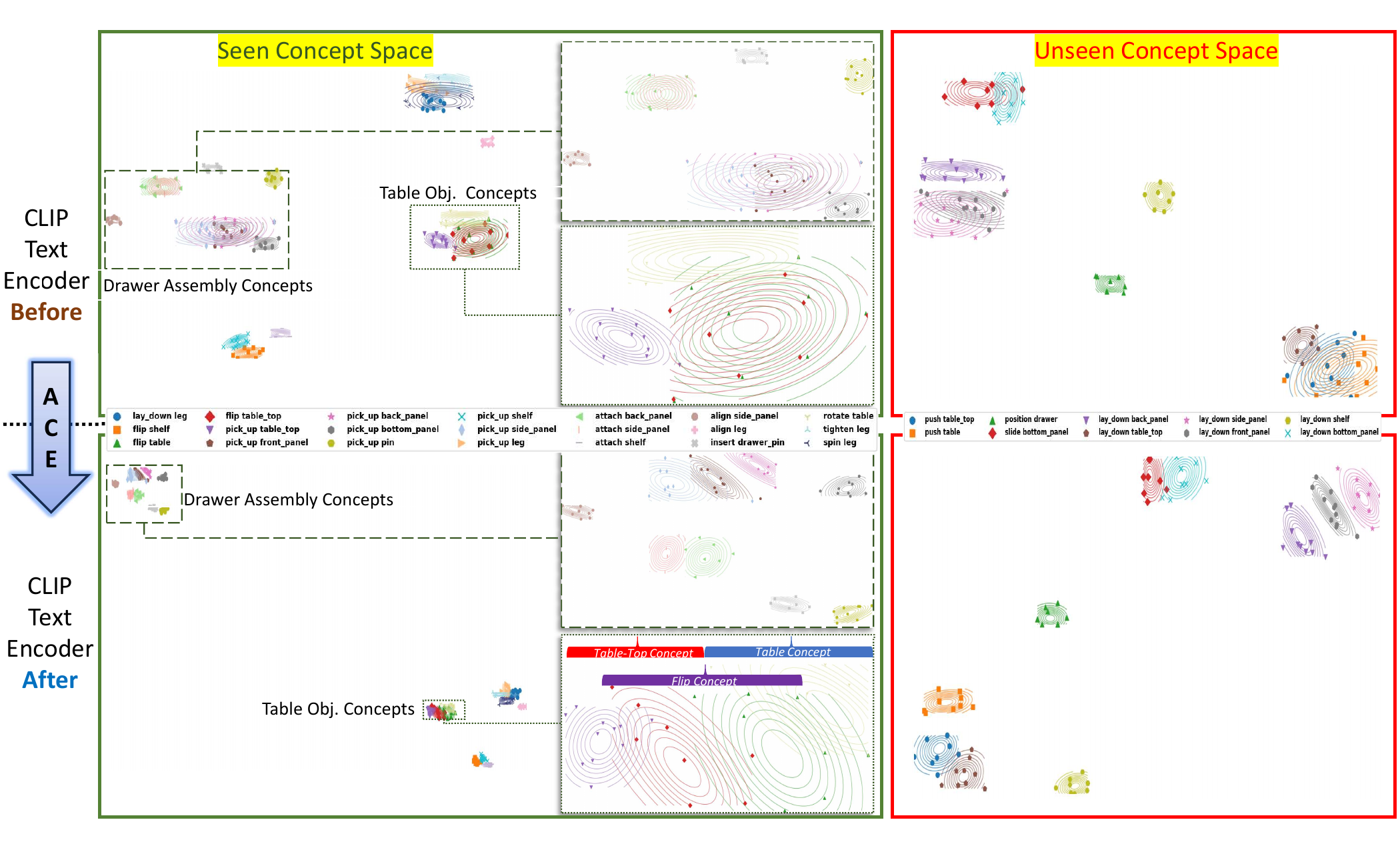}
\caption{TSNE visualization for synonym embeddings in seen (outlined in green) and unseen (outlined in red) action spaces of IKEA dataset. The original CLIP embeddings of action synonyms are ACEd and grouped more distinctly. Please zoom in to see finer details.}
\label{fig:tsne}
\end{figure*}
\subsection{Concept Space Visualization}
Fig. \ref{fig:tsne} shows the TSNE visualization of action concept subspaces in the IKEA dataset. Each action subspace is estimated by a Gaussian resembling a galaxy, containing its synonym embeddings (planets) encoded by the text encoder. The top and bottom halves of Fig.\ref{fig:tsne} compare synonym embeddings from the original CLIP (trained on 400M web image-text pairs) and ACEd CLIP for both base and novel action classes. For base classes, only two representative synonyms were seen during training, with the rest unknown to ACEd CLIP. Zoomed-in views of synonym embeddings for drawer and table-top classes are provided.

When comparing ACEd CLIP with the original CLIP in both seen and unseen spaces, ACEd embeddings have a smaller standard deviation, with action synonyms aligned more closely. This increases distinction between subspaces of different action classes and makes the model more invariant to synonyms. For example, while \textit{push table}'', \textit{push table-top}'', and \textit{lay down table-top}'' overlap in original CLIP, their ACEd embeddings are grouped more distinctly. This is notable since none of these concepts or their synonyms were seen during training. Additionally, ACEd CLIP better aligns similar concepts, unlike original CLIP, which lacks hierarchical understanding. In the seen concept space, embeddings for the drawer and table assemblies are close, while ACEd embeddings for different furniture assemblies are mapped far apart. Similarly, in the unseen concept space, synonyms of \textit{lay down shelf}'' initially project between drawer and table concepts but shift toward table assemblies after ACE, even though ``shelf'' is not used in other unseen action classes.

\section{Conclusion}
This paper focused on recognizing unseen procedural steps in trimmed videos. We demonstrated that current vision-language models lack intrinsic action concept understanding and tend to overfit to fixed labels. To address this, we introduced Action Concept Enhancement (ACE), a fine-tuning technique that improves VLMs' robustness and conceptual understanding. By integrating action synonyms stochastically during training, ACE mitigates overfitting to fixed verbs and objects. Our experiments in the cooking and assembly domains show that ACE significantly enhances zero-shot action concept recognition while maintaining competitive performance on seen actions.

\section{Appendix}

In this appendix, we provide important implementation details of our method as well as the list of unseen synonyms used during Synonym Robustness Test (SRT). We also compare our model with LaVila\cite{lavilla}, as an egocentric-focued baseline, on egocentric videos of the GTEA\cite{gtea} dataset.
\subsection{Implementation Details}

We use a 12-layer TimeSformer \cite{timesformer} video encoder pretrained on Howto100M via ProcVLR \cite{proceVLR} and the original 12-layer CLIP text encoder (ViT-B/16). The video input to our encoder comprises of 30 frames over  3 seconds  in ATA, 25 frames over  2 seconds  in IKEA, and 15 frames over 1 second in GTEA videos. These temporal windows are centered at the mid point of each action segment during training. For evaluation, we follow the practice of sampling three temporal clips of 224x224 crops per video and report the average \cite{slowfast}. 
GPT-4 \cite{gpt4} generates synonyms for our method.  The number of first and second order children in synonym trees are 2,9 and 11 for the IKEA, GTEA and ATA datasets, respectively. Furthermore, batch size is set to 16, temperature $\tau$ is adjusted to 0.02, and SGD optimizes the model for up to 15 epochs on ATA, and 12 epochs for IKEA and GTEA datasets.  Our parameters are the same for our VLM and the one used in our direct baseline ProcVLM \cite{proceVLR}.

Importantly, during training at each iteration, we ensure that actions sharing the same root verb are assigned the same randomly-selected verb synonym to compute $L_{rand}$. However, during testing, actions with the same root verb may have different synonyms chosen in each run, as synonyms for each action are sampled independently at test time (refer to Table \ref{table:ATA_SRT}-\ref{table:GTEA_SRT}).

Furthermore, we prompt GPT-4 as follows to generate $M$ synonyms for action X of the ATA dataset as an example:

\noindent \textit{"what are M synonyms of the action (X) during toy assembly? please follow the constraints below: }

          \noindent\textit{1- list each synonym in a new line without any numbering and  period or commas.}
          
          \noindent\textit{2- make sure the resulting sentences semantically and contextually make sense given the assembly context.}
          
          \noindent\textit{3-  start each line with a verb and all in small letters.}
          
          \noindent\textit{4- use the same object and sentence structure as the query.}
          
          \noindent\textit{5- if the verb is a phrasal verb like 'put down', then place the whole phrasal verb in the begging of the sentence. }
          
          \noindent\textit{6- if the query is a phrasal verb, then I encourage you to output phrasal verbs too, specially if the phrasal verb indicates some spatial information about the scene.}
        "

 \subsection{Comparison with LaVila}
 LaVila is a video-language model that is pretrained on augmented captions in long-term and untrimmed videos. Having said that, LaVila's pretrained checkpoints are only available based on the egocentric videos of Ego4D\cite{ego4d}. On the other hand, ACE is pretrained on execontric videos of Howto100M. Therefore, a direct comparison between LaVila and ACE on GTEA's egocentric videos is not entirely fair, as LaVila has a pretraining advantage. Nonetheless, we compare our method with LaVila in Table \ref{table:comparison_gtea}. While LaVila outperforms ACE on default labels, it struggles with action synonym understanding, where ACE demonstrates more robust performance, despite being pretrained on third-person videos.

\begin{table*}[t]
\begin{center}
\footnotesize\setlength{\tabcolsep}{1 pt}
\caption {Action classification \textit{acc} of the egocentric GTEA videos.\label{table:comparison_gtea}} 
\begin{tabular}{c|c|c c c|c}
\hline
\multicolumn{2}{c|}{}& 
\multicolumn{4}{|c}{GTEA Dataset \textit{[6 base and 4 novel classes]}}   \\ 
\cline{3-6}
 \multicolumn{2}{c|}{} & \multicolumn{3}{|c|}{Default Labels} & SRT \\ 
\hline
 Method &Pretrained on& Seen & Unseen & HM & Unseen \\ 
\hline
LaVila \cite{lavilla} & Ego4D\cite{ego4d} &\cellcolor{gray!20}\textbf{96.0} & \textbf{71.4} & \cellcolor{gray!20}\textbf{81.9} &37.5$\pm$21.0 \\
\hline
\textbf{Ours}&Howto100M\cite{howto100m}  & \cellcolor{gray!20}85.1 & 67.2&\cellcolor{gray!20}75.1& \textbf{45.0$\pm$16.8}\\ \hline
\end{tabular}
\end{center}
\end{table*}

\subsection{Synonym Robustness Test (SRT) Labels}
 In order to make our SRT results comparable with future work, we provide the GPT-generated synonyms for the novel actions across our three benchmark datasets (Table \ref{table:ATA_SRT}-\ref{table:GTEA_SRT}). The default/root labels are indicated in bold in each table. In few instances, for a given run, the connection between the AI-generated synonyms and their underlying action concept might be loose, yet such synonyms describe their associated concepts more accurately relative to the labels of other actions in the same run. Besides, the coarser labels evaluate the action concept understanding of VLMs at various degrees of granularity. A robust method should have a high mean a low standard deviation when tested against these 10 sets of action synonyms.

\begin{table*}[t]
\begin{center}
\footnotesize\setlength{\tabcolsep}{3.0 pt}
\caption {GPT-genrated action synonyms for the Synonym Robustnss Test (SRT) on the novel classes of the ATA dataset. Labels in each column correspond to the same action concept. The root labels are highlighted in bold.} \label{table:ATA_SRT}
\begin{tabular}{c|c|c|c | c|c}
\hline
 \multicolumn{1}{c|}{Run} & \multicolumn{5}{c}{Unseen Action Labels - ATA } \\ 
\hline
\textbf{1} & \textbf{'drop item'} & \textbf{'balance part'}& \textbf{'pick up item'}& \textbf{'spin block'} & \textbf{'hammer pin'} \\
2 & 'leave item'& 'stabilize part'& 'clutch item'& 'wheel block'& 'whack pin' \\
3 & 'set down item'& 'center part'& 'retrieve item'& 'turn block'& 'nail pin' \\
4 & 'lower item'&'align part'& 'grab item'& 'twirl block'& 'pound pin' \\
5 & 'let fall item'& 'steady part'& 'catch item'& 'circle block'& 'bash pin' \\
6 & 'deposit item'& 'level part'& 'hold item'& 'swivel block'& 'beat pin' \\
7 & 'release item' & 'adjust part'& 'lift item'& 'rotate block'& 'drive pin' \\
8 & 'place item'& 'calibrate part'& 'take item'& 'revolve block'& 'strike pin' \\
9 & 'lay down item'& 'match part'& 'grasp item'& 'twist block'&'thunk pin' \\
10 & 'ditch item'& 'weigh part'& 'harness item'& 'wind block'& 'smash pin' \\
\end{tabular}
\end{center}
\end{table*}

\begin{table*}[h!]
\begin{center}
\tiny\setlength{\tabcolsep}{0.5 pt}
\caption {GPT-genrated action synonyms for the Synonym Robustnss Test (SRT) on the novel classes of the IKEA dataset. Labels in each column correspond to the same action concept. The root labels are highlighted in bold. Please zoom in for a better view.} \label{table:IKEA_SRT}
\begin{tabular}{c|c|c|c|c |c |c |c |c |c | c}
\hline
 \multicolumn{1}{c|}{Run} & \multicolumn{10}{c}{Unseen Action Labels - IKEA } \\ 
\hline
\textbf{1} & \textbf{'lay down shelf'}& \textbf{'lay down side panel'}& \textbf{'lay down front panel'}&\textbf{'push table top'}& \textbf{'position drawer right side up'}& \textbf{'lay down table top'}& \textbf{'slide bottom panel'}& \textbf{'push table'}& \textbf{'lay down bottom panel'}& \textbf{'lay down back panel'}  \\
2 & 'set down shelf'& 'place side panel'& 'set down front panel'& 'press table top'& 'place drawer right side up'& 'put down table top'& 'insert bottom panel'& 'press table'& 'set down bottom panel'& 'place down back panel'\\
3 & 'position down shelf'& 'rest side panel'& 'position down front panel'& 'shift table top'& 'align drawer right side up'& 'position down table top'& 'push bottom panel'& 'shove table'& 'position down bottom panel'& 'put down back panel'\\
4 & 'put down shelf'& 'position side panel'& 'place down front panel'& 'slide table top'& 'set drawer right side up'& 'set down table top'& 'guide bottom panel'& 'slide table'& 'put down bottom panel'& 'set down back panel'\\
5 & 'position down shelf'& 'rest side panel'& 'place down front panel'& 'push table top'& 'orient drawer right side up'& 'set down table top'& 'guide bottom panel'& 'press table'& 'place down bottom panel'& 'set down back panel'\\
6 &'set down shelf'& 'rest side panel'& 'lay down front panel'& 'nudge table top'& 'orient drawer right side up'& 'position down table top'& 'move bottom panel'& 'shove table'& 'position down bottom panel'& 'set down back panel'\\
7 & 'place down shelf'& 'place side panel'& 'put down front panel'& 'nudge table top'& 'orient drawer right side up'& 'put down table top'& 'move bottom panel'& 'nudge table'& 'put down bottom panel'& 'place down back panel'\\
8 & 'put down shelf'& 'lay down side panel'& 'set down front panel'& 'nudge table top'& 'align drawer right side up'& 'place down table top'& 'guide bottom panel'&'press table'& 'place down bottom panel'&'position down back panel'\\
9 & 'position down shelf'& 'rest side panel'& 'put down front panel'&'nudge table top'& 'align drawer right side up'& 'set down table top'& 'slide bottom panel'& 'shove table'& 'position down bottom panel'& 'place down back panel'\\
10 & 'position down shelf'& 'set down side panel'& 'lay down front panel'& 'slide table top'& 'align drawer right side up'& 'set down table top'& 'guide bottom panel'& 'push table'& 'lay down bottom panel'& 'put down back panel'\\
\end{tabular}
\end{center}
\end{table*}

\begin{table*}[h!]
\begin{center}
\footnotesize\setlength{\tabcolsep}{3.0 pt}
\caption {GPT-genrated action synonyms for the Synonym Robustnss Test (SRT) on the novel classes of the GTEA dataset. Labels in each column correspond to the same action concept. The root labels are highlighted in bold. Given that we only use verbs for the GTEA datasets,  the term ``ingredient'' is added to verbs as a placeholder to comply with the verb+object template of the actions.} \label{table:GTEA_SRT}
\begin{tabular}{c|c|c|c | c}
\hline
 \multicolumn{1}{c|}{Run} & \multicolumn{4}{c}{Unseen Action Labels - GTEA } \\ 
\hline
\textbf{1} & \textbf{'shake ingredient'} & \textbf{'fold ingredient'}& \textbf{'stir ingredient'}& \textbf{'spread ingredient'}  \\
2 & 'toss ingredient'& 'place together ingredient'& 'mix ingredient'& 'smear ingredient'\\
3 & 'rattle ingredient'& 'combine ingredient'& 'beat ingredient'& 'smooth ingredient'\\
4 & 'jostle ingredient'& 'tuck ingredient'& 'whisk ingredient'&'garnish ingredient'\\
5 & 'tremble ingredient'& 'incorporate ingredient'& 'whip ingredient'& 'slather ingredient'\\
6 & 'sway ingredient'& 'integrate ingredient'& 'swirl ingredient'& 'apply ingredient'\\
7 & 'vibrate ingredient'& 'mingle ingredient'& 'rotate ingredient'& 'cover ingredient'\\
8 & 'rock ingredient'& 'merge ingredient'& 'agitate ingredient'& 'layer ingredient'\\
9 & 'quiver ingredient'& 'interlace ingredient'& 'incorporate ingredient'& 'lather ingredient'\\
10 & 'wobble ingredient'& 'fuse ingredient'& 'dissolve ingredient'& 'daub ingredient'\\
\end{tabular}
\end{center}
\end{table*}

\newpage

{\small
\bibliographystyle{ieee_fullname}
\bibliography{egbib}
}

\end{document}